\documentclass{article}

\usepackage{microtype}
\usepackage{graphicx}
\usepackage{caption}
\usepackage{subcaption}
\usepackage{booktabs} 
\usepackage{amsmath}
\usepackage{amssymb}
\usepackage{tabularx}
\usepackage[ruled,linesnumbered]{algorithm2e}
\usepackage{url}

\usepackage[accepted]{icml2019}

\icmltitlerunning{Sparse evolutionary deep learning with over one million artificial neurons on commodity hardware}

\begin{document}

\twocolumn[
\icmltitle{Sparse evolutionary Deep Learning with over one million artificial neurons on commodity hardware}




\begin{icmlauthorlist}
\icmlauthor{Shiwei Liu}{tue}
\icmlauthor{Decebal Constantin Mocanu}{tue,twente}
\icmlauthor{Amarsagar Reddy Ramapuram Matavalam}{Iowa}
\icmlauthor{Yulong Pei}{tue}
\icmlauthor{Mykola Pechenizkiy}{tue}
\end{icmlauthorlist}

\icmlaffiliation{tue}{Department of Mathematics and Computer Science,
              Eindhoven University of Technology, 5600 MB Eindhoven, the Netherlands\\}
\icmlaffiliation{twente}{Faculty of Electrical Engineering, Mathematics andComputer Science, University of Twente, Enschede 7522NB,The Netherlands\\}
\icmlaffiliation{Iowa}{Department of Electrical and Computer Engineering, Iowa State University, USA\\}
\icmlcorrespondingauthor{Shiwei Liu}{s.liu3@tue.nl }

\icmlkeywords{scalable deep learning, adaptive sparse connectivity, sparse evolutionary training, curse of dimensionality, microarray gene expression, high-dimensional data }

\vskip 0.3in
]



\printAffiliationsAndNotice{}  

\begin{abstract}
Artificial Neural Networks (ANNs) have emerged as hot topics in the research community. Despite the success of ANNs, it is challenging to train and deploy modern ANNs on commodity hardware due to the ever-increasing model size and the unprecedented growth in the data volumes. Particularly for microarray data, the very-high dimensionality and the small number of samples make it difficult for machine learning techniques to handle. Furthermore, specialized hardware such as Graphics Processing Unit (GPU) is expensive. Sparse neural networks are the leading approaches to address these challenges. However, off-the-shelf sparsity inducing techniques either operate from a pre-trained model or enforce the sparse structure via binary masks. The training efficiency of sparse neural networks cannot be obtained practically. In this paper, we introduce a technique allowing us to train truly sparse neural networks with fixed parameter count throughout training. Our experimental results demonstrate that our method can be applied directly to handle high dimensional data, while achieving higher accuracy than the traditional two phases approaches. 
Moreover, we have been able to create truly sparse MultiLayer Perceptrons (MLPs) models with over one million neurons and to train them on a typical laptop without GPU ( \url{https://github.com/dcmocanu/sparse-evolutionary-artificial-neural-networks/tree/master/SET-MLP-Sparse-Python-Data-Structures}), this being way beyond what is possible with any state-of-the-art techniques.
\icmlkeywords{truly sparse neural networks, sparse evolutionary training (SET), microarray gene expression, adaptive sparse connectivity }

\end{abstract}

\section{Introduction}
\label{intro}
In the past decades, Artificial Neural Networks (ANNs) have become an active area of current research due to state-of-the-art performance they have achieved in a variety of domains, including image recognition, text classification, and speech recognition. The powerful hardware, like Graphics Processing Unit (GPU), as well as the increasing growth of data volumes, accelerate the advances of ANNs significantly. Recently, some works \cite{belkin2019reconciling,elton2020self} show that increasing the model capacity beyond a particular threshold yields better generalization. However, GPU is expensive and the explosive increase of model size leads to prohibitive memory requirements. Thus, the required resources to train and employ the modern ANNs are at odds with commodity hardware where the resources are very limited.

Motivated by these challenges, sparse neural networks \cite{lecun1990optimal,chauvin1989back} have been introduced to effectively reduce the memory requirements to deploy ANN models. After that, various techniques have emerged to obtain sparse neural networks, including but not limited to pruning  \cite{han2015learning,narang2017exploring,h.2018to,frankle2018lottery}, $L_0$ and $L_1$ Regularization \cite{louizos2017learning,wen2017learning}, Variational Dropout \cite{molchanov2017variational}, soft weight-sharing \cite{ullrich2017soft}. While achieving a high level of sparsity and preserving competitive performance, these methods usually involve a pre-trained model and a re-training process, which makes the training process remain inefficient.

Recently, several works have developed techniques allowing to train sparse neural networks with fixed parameter budget throughout the training based on adaptive sparse connectivity, e.g. Sparse Evolutionary Training (SET) \cite{mocanu2018scalable}, DEEP-R \cite{bellec2017deep}, Dynamic Sparse Reparameterization (DSR) \cite{mostafa2019parameter}, Sparse Momentum \cite{dettmers2019sparse}, ST-RNNs \cite{liu2019intrinsically}, Rigged Lottery (RigL)\cite{evci2020rigging}. The sparse weights, initialized with a fixed sparsity (a fraction of model parameters with zero values), can be maintained throughout training. The heuristic behind these techniques is following a cycle of weight pruning and weight regrowing based on a certain criterion. Essentially, the whole process of sparse training can be treated as a combinatorial optimization problem (weights and sparse structures). As the number of parameters during training is strictly constrained, sparse training techniques based on adaptive sparse connectivity are able to achieve the training efficiency as well as the inference efficiency associated with the final compressed model. However, due to the limited support for sparse operations in GPU-accelerated libraries, the sparse structure is enforced with binary masks. Thus, the training efficiency is only demonstrated theoretically not practically.

Due to the above-mentioned problems, the memory requirements and computation capacity to directly train wide neural networks with hundreds of thousands of neurons to deal with high dimensional non-spatial like data (e.g., tabular data) with over 20,000 dimensions(input features) and less than 100 samples, are usually beyond what is allowed on commodity hardware. This paper aims to process high-dimensional data with a truly sparse end-to-end model. More precisely, we focus on the original SET algorithm because it was shown that it is capable of reaching very high accuracy performance \cite{zhu2019multi, mocanu2018scalable}, many times even higher than the dense counterparts \cite{liu2019intrinsically}, while being very versatile and suitable for many neural network models (e.g. restricted Boltzmann machines \cite{mocanu2017estimating}, multilayer perceptrons \cite{liu2019improving}, and convolutional neural networks) and non-grid like data. However, due to the limitations of typical deep learning libraries (e.g. optimized operations just for fully-connected layers and dense matrices), the largest number of neurons used in \cite{mocanu2018scalable} is just 12,082 neurons - quite a low representational power. Practically, the original SET-MLP implementation uses the typical approach
from the literature to work with sparsely connected layers,
i.e. fully connected layers with sparsity enforced by a binary
mask over their weights - this approach, of course, is far
from using the full advantage of sparsity. Instead of generating a mask to enforce sparsity, in this paper, we devise the first sparse implementation for adaptive sparse connectivity so that it is possible to design neural network models which are very large in terms of representational power, but small in terms of space complexity to fit onto memory-limited devices.

The first contribution of this paper is a truly sparse implementation of SET, which can create and train SET-MLP models with hundreds of thousands of neurons on a typical laptop without GPU to handle data with tens of thousands of dimensions, a situation which is over the capacity of traditional fully-connected MLPs. Secondly, we show that our proposed approach can be a good replacement for the current methods which employ both, feature reduction and classifiers, to perform classification on high-dimensional non-image datasets. Thirdly, we show that our proposed solution is robust to the  ``curse of dimensionality", avoiding overfitting and achieving very good performance in terms of classification accuracy on a dataset with over 20,000 dimensions (input features) and less than 100 samples.

\section{Related work}
\label{sec:1}
In this section, we will introduce the advances of processing high-dimensional microarray data and techniques allowing training sparse neural networks from scratch.
\subsection{Artificial neural networks on microarray gene expression}
\label{sec:2}

Data have become indispensable factors of the success of machine learning (ML). The performance of a ML application is primarily determined by the quality and the quantity of the training data \cite{lecun2015deep}.
Especially, gene expression obtained from DNA microarray has emerged as a powerful solution to cancer detection and treatment \cite{simon2003pitfalls}. However, most of the datasets in DNA microarray are high-dimensional and redundant, which would result in the unnecessary calculation, large memory requirement and even the decrease of generalization ability due to the ``curse of dimensionality''\cite{destrero2009feature}. Moreover, the invisible relationships and non-standard structures among different features also make it very time-consuming to find the key features from tens of thousands of features. 

To tackle this problem, various methods have been proposed by researchers. Among them, feature selection is undoubtedly a ``de facto'' standard as it is not only able to remove the redundant features and to keep the important ones, but it also helps to improve the model performance\cite{destrero2009feature}. Following the feature detection phase, standard machine learning classifiers can be used to perform classification on the selected features. Traditional feature selection methods can be roughly divided into three categories: filter methods \cite{guyon2003introduction, forman2003extensive,bekkerman2003distributional,caruana2003benefitting,koller1996toward,davidson2010feature,de2019fast,de2009sofmls}, wrapper methods \cite{kohavi1997wrappers,pudil1994floating,reunanen2003overfitting,eshelman1991chc, cordon2006feature,meda2018estimation,de2017usnfis} and embedded methods \cite{langley1994selection, mundra2009svm, romero2008performing}. 
Independent of the classifier, filter methods are able to deal with large scale datasets efficiently as they have low computational costs due to the fact that they select variables using proxy measures (e.g. mutual information), not an error metric provided by the classifier \cite{pohjalainen2015feature}.
Wrapper methods employ feedback classification accuracy to assess the different suboptimal subsets chosen by following search algorithms, which can have good results but also increases the computation cost\cite{chandrashekar2014survey}. The WrapperSubsetEval\cite{hall2009weka} is a general wrapper method which can be connected with various learning algorithms. Different from the previous two discussed categories, in embedded methods \cite{mejia2006feature}, the feature selection and the classifier are not separated from each other. 

As more and more datasets with ultra-high dimensions have emerged, these datasets also bring challenges to conventional algorithms running on normal computers due to the expensive computational costs. To address this problem, distributed computing has been proposed. A distributed decentralized algorithm for \textit{k}-Nearest Neighbor (kNN) graph has been proposed in ~\cite{plaku2007distributed}. This framework is able to distribute the computation of the kNN graph with very big datasets by utilizing the sequential structure of kNN data. MapReduce~\cite{dean2008mapreduce} is an efficient programming model 
used by Google to compute different types of data and process large raw data. 
Moreover, a classifier framework combining MapReduce and proximal Support Vector Machine(mrPSVM) has been proposed in~\cite{kumar2015classification}. The results on several high-dimensional, low-sample benchmark datasets demonstrate that the ensemble of mrPSVM classifier with feature selection methods using statistical tests outperforms classical approaches. 

Although the above-mentioned hierarchical algorithms can have a good performance on classification tasks, the proper performance heavily depends on the features meticulously selected by experts from different domains \cite{min2017deep}. This means that at least a dimensionality reduction technique is needed before the classifier. As an emerging branch of machine learning, deep neural networks tackle this problem via the explosive increase of data and computation ability. Multi-Layer Perceptron (MLP) is one of the most used architectures in deep neural networks, e.g. it represents 61\% of a typical Google TPU (Tensor Processing Unit) workload for production neural networks applications, while convolutional neural networks represent just 5\% \cite{jouppi2017datacenter}. However, it is difficult to employ MLPs directly on high dimensional data tasks due to the quadratic number of parameters in its fully-connected layers. This limits MLPs size to several thousand neurons and a few thousand input features on commodity hardware, and implicitly their representational power.

Being a successful approach that has been widely used in image recognition, speech recognition, language translation, etc., deep neural networks have also been employed to deal with high-dimensional data.
MLPs have been widely applied to solve gene expression regulation problems. Chen et al. \cite{chen2016gene} have presented MLPs for gene expression inference (D-GEX) to perform gene expression inference to the GEO microarray data and RNA-seq expression data. An autoencoder has been connected with principal component analysis (PCA) to learn the high-level features of 13 microarray data \cite{fakoor2013using}. Convolutional neural networks (CNNs) are also used to solve biological sequence problems due to its outstanding capability to learn spatial information. Alipanahi et al. \cite{alipanahi2015predicting} have proposed a CNN-based approach, called DeepBind, to handles both microarray and sequencing data. By two downstream applications, DeepBind can automatically analyze sequencing data and alleviate the time-consuming human designing work. 
\subsection{Intrinsically sparse neural networks}
\label{sec:3}
Recently, there are some works attempting to train an intrinsically sparse neural network from scratch to obtain the efficiency both for the training and inference phases. Mocanu et al. \cite{mocanu2016topological} have trained sparse restricted Boltzmann machines that have fixed scale-free and small-world connectivity. After that, \cite{mocanu2017network, mocanu2018scalable} have introduced the Sparse Evolutionary Training procedure and the concept of adaptive connectivity for intrinsically sparse networks to fit the data distribution. The Nest algorithm \cite{dai2019nest} gets rid of a fully-connected network at the beginning by a grow-and-prune paradigm, that is, expanding a small randomly initialized sparse network to a large one and then shrink it down. A Bayesian posterior has been applied to sample the sparse network configurations, while providing a theoretical guarantee for connectivity rewire \cite{bellec2017deep}. Besides weights pruning and regrowth, cross-layer weights redistribution has been used to adjust network architectures for better performance \cite{mostafa2019parameter}. Liu et al. \cite{liu2019improving} have further reduced the number of parameters by applying neurons pruning, while getting competitive performance. Dettmers et al. \cite{dettmers2019sparse} have used the momentum information of momentum Stochastic gradient descent to tackle weights regrowth and redistribution problems, reaching dense performance levels with 35-50\%, 5-10\%, and 20-30\% weights for AlexNet, VGG16 and Wide Residual Networks, respectively. Very recently, by modifying the sparsity distribution of \textit{Erd{\H{o}}s-R{\'e}nyi} introduced in \cite{mocanu2018scalable}, RigL \cite{evci2020rigging} can match and sometimes exceed the performance of pruning based approaches. On the other hand, the Lottery Ticket Hypothesis has been proposed to find the sparse networks that can reach better accuracy \cite{frankle2018lottery} than dense networks. However, a dense network trained at the beginning limits its efficiency only for inference, not the training process. While achieving proper performance, these methods demonstrated computational efficiency via applying a binary mask on the weights due to the lack of efficient sparse linear algebra support from processors like TPUs or GPUs. 
\subsection{Sparse evolutionary training}
\label{sec:4}
Inspired by the fact that biological neural networks are prone to be sparse, rather than dense \cite{strogatz2001exploring, pessoa2014understanding}, there is an increasing interest in conceiving neural networks with a sparse topology \cite{mocanu2016topological,yoon2017lifelong}. 
In \cite{mocanu2018scalable}, the authors proposed a novel concept, sparse neural networks with adaptive sparse connectivity to maintain sparsity during training. Given a dataset $\mathbf{D} = \{(x_i, y_i) \}_{i=1}^n$, let a network denoted by:
\begin{equation}
\hat{y} = f(x;\theta)
\end{equation}
where the $f(x_i;\theta)$ is the neural network parameterized by $\mathbf{\theta}$. The parameters $\theta$ can be decomposed into dense matrix $\theta^l \in {\mathbf{R}}^{n^{l-1}\times{n^l}}$, where $n^l$ and $n^{l-1}$ represent the number of neurons of the layer $l$ and $l-1$, respectively. We train the network to minimize the loss function $\sum L(f(x;\theta),y)$. The motivation of sparse neural networks is to reparameterize the dense network only with a fraction of parameters, $\theta_s$. The parameters $\theta_s$ can be decomposed into sparse matrix $\theta^l_s \in {\mathbf{R}}^{n^{l-1}\times{n^l}}$, for each layer $l$. A sparse neural network can be demoted by:
\begin{equation}
\hat{y}_s = f_s(x;\theta_s)
\end{equation}
Let us define the sparsity of the network as $S = 1 - \frac{\|\theta_s\|_0}{\|\theta\|_0}$, where $\| \theta \|_0$ refers to the $l_0$ norm of $\theta$.

The sparse evolutionary training (SET) is a method that allows efficiently training sparse neural networks from scratch with a fixed number of parameters. The basic idea underlying SET is first initializing a network with a sparse topology and then optimizing the weight values and the sparse topology together during the training process, to fit the data distribution. Different from the conventional methods, e.g. weights pruning \cite{Cun90optimalbrain, Han:2015:LBW:2969239.2969366} which creates sparse topologies during or after the training process, the network trained with SET is designed to be sparse before training. This quadratically reduces the number of connections during the whole training phase. The main parts of SET are sparse initialization and the weight pruning-regrowing cycles, explained below.


\textbf{(1) Sparse initialization} The initial sparse topology proposed in SET is  \textit{Erd{\H{o}}s-R{\'e}nyi} random graph topology~\cite{24gotErdos1959} where a sparse matrix $\theta_s^l \in {\mathbf{R}}^{n^{l-1}\times{n^l}}$ represents connections between two consecutive layers $l-1$ and $l$. More precisely, the network is initialized by:
\begin{equation}
 \theta_s^l = \theta^l * M^l
 \label{Eq:initialization}
\end{equation}
where $*$ represents the Hadamard product and $M^l$ is a binary matrix of the same size with $\theta^l$, in which each element $M^l_{i,j}$ is given by the probability $P(M_{i,j}^l)=\min(\frac{\epsilon(n^l+n^{l-1})}{n^l\times{n^{l-1}}},1)$. $\epsilon\in\mathbf{R^+}$ is a hyperparameter to control the sparsity level $S$. Such initialization distributes higher sparsity to the layers where $n^l$ is approximately in the same range with $n^{l-1}$, and  lower sparsity to the layers where $n^l\gg n^{l-1}$ or vice-versa.

\textbf{(2) Weight pruning-regrowing cycle}
After each training epoch, unimportant connections (accounting for a a certain fraction $\zeta$ of $||M^l||_0$) will be pruned in each layer. The remaining connections are given by:
\begin{equation}
 \theta_s^l = \theta_s^l * (M^l - P^l)
 \label{Eq:pruning}
\end{equation}
where $P^l$ is a binary matrix with the same size as $M^l$, $||P^l||_0=\zeta||M^l||_0$, and the non-zero elements of $P^l$ is a subset of the non-zero elements of $M^l$ corresponding to largest negative weights and the smallest positive weights in $\theta^l_s$. After that, an equal number of connections with $\zeta||M^l||_0$ are randomly added to each layer by:
\begin{equation}
 \theta_s^l = \theta_s^l + \theta^l_r
 \label{Eq:regrowing}
\end{equation}
where $\theta^l_r \in {\mathbf{R}}^{n^{l-1}\times{n^l}}$ has exactly $\zeta||M^l||_0$ non-zero values. The non-zero element locations from $\theta^l_r$ are picked using a random uniform distribution, and their values are set using a small Gaussian noise. Finally, $M^l$ is updated as follows:
\begin{equation}
    M^l_{i,j}=\begin{cases}
    1, & \text{if $\theta_s^l(i,j)$ $\neq$ $0$}\\
    0, & \text{otherwise}.
  \end{cases} \text{ }\forall i,j
\end{equation}
Roughly speaking, the removal of the connection in SET represents natural selection, whereas the emergence of new connections corresponds to the mutation phase in natural evolution inspiring computing. 

However, the authors of SET have used Keras with Tensorflow backend to implement their SET-MLP models. This implementation choice, while having the significant advantage of offering wide flexibility of architectural choices (e.g. various activation functions, optimizers, GPUs, and so on), which is very welcomed while conceiving new algorithms, does not offer proper support for sparse matrix operations. This limits the practical aspects of SET-MLP considerably with respect to its maximum possible number of neurons and implicitly to its representational power. Due to these reasons, the largest SET-MLP model reported in the original paper \cite{mocanu2018scalable} only contains 12,082 neurons on 
NVIDIA Tesla M40. Note that it is possible to increase the size of such SET-MLP implementations with several thousands more neurons, but no chance to reach one million neurons.

\begin{algorithm}
\DontPrintSemicolon
\KwData{A neural network $M$ with $L$ layers, Weight $\theta$, sparsity $S
$, pruning rate $\zeta$, training epoch $n$}
\% Sparse initialization \\
\For{$l\leftarrow 0$ \KwTo $L$}{
$\theta_s^l = \theta^l * M^l$
}
\% Weight pruning-regrowing cycle \\
\For{$epoch\leftarrow 0$ \KwTo $n$}{
$\theta_s\leftarrow$ NormalTraining$(\theta_s)$ \\
\For{$l\leftarrow 0$ \KwTo $L$}{
$\theta_s^l\leftarrow$ $\theta_s^l * (M^l - P^l)$ \\
$\theta_s^l\leftarrow$ $\theta_s^l * (M^l + R^l)$ \\
}
}
\caption{Sparse evolutionary training (SET)}
\end{algorithm}

\section{Proposed method}
\label{sec:5}
In this paper, we address the above limitations of the SET original implementation and we show how vanilla SET-MLP can be implemented from scratch using just pure Python, SciPy, and Cython. Our approach enables the construction of SET-MLPs with at least two orders of magnitude larger, i.e. over 1,000,000 neurons. What is more, such SET-MLPs do not need GPUs and can run perfectly fine on a standard laptop. 

\subsection{Sparse matrices operations}
\label{sec:6}
The key element of our very efficient implementation is to use sparse data structures from SciPy. It is important to use the right representation of a sparse matrix for different operations because different sparse matrix formats have different advantages and disadvantages. Below the SciPy sparse data structures used to implement SET-MLPs are briefly discussed, while the interested reader is referred to\footnote{\url{https://docs.scipy.org/doc/scipy/reference/sparse.html.} Last visit 3rd June 2018.} for detailed information.
\begin{itemize}
\item Compressed Sparse Row (CSR) sparse matrix: The data is stored in three vectors. The first vector contains nonzero values, the second one stores the extents of rows, and the third one contains the column indices of the nonzero values. This format is very fast for many arithmetic operations but slow for changes to the sparsity pattern.
\item Linked List (LIL) sparse matrix: This format saves nonzero values in row-based linked lists. Items in the rows are also sorted. The format is fast and flexible in changing the sparsity patterns but inefficient for arithmetic matrix operations.
\item Coordinate list (COO) sparse matrix: This format saves the nonzero elements and their coordinates (i.e. row and column). It is very fast in constructing new sparse matrices, but it does not support arithmetic matrix operations and slicing.
\item Dictionary Of Keys (DOK) sparse matrix: This format has a dictionary that maps row and column pairs to the value of nonzero elements. It is very fast in incrementally constructing new sparse matrices, but can not handle arithmetic matrix operations.
\end{itemize}
 Note that, one format cannot handle all operations necessary for sparse weights matrices to implement a SET-MLP. Still, the conversions from one format to another are very fast and efficient. Thus, in our implementation which was done in pure Python 3, we have used for specific SET-MLP operations, specific sparse matrix formats and their fast conversion capabilities, as follows. 

\textit{Initialize sparsely connected layers}. The sparse matrices which store the sparsely connected layers are creating using the Linked List (LIL) format and then are transformed into Compressed Sparse Row (CSR) format.

\textit{Feed-forward phase.} During it, the sparse weights matrices are stored and used in the CSR format.

\textit{Backpropagation phase - computing gradients.} The only operations which can not be implemented with SciPy sparse matrix operations is computing the gradients for backpropagation~\cite{rumelhart1986learning} due to the simple fact that by multiplying the vector of backpropagation errors from layer $h^k$ with the vector of activation neurons from layer $h^{k-1}$ will perform a considerable amount of unnecessary multiplications (for nonexistent connections) and will create a dense matrix for updates. This dense matrix, besides being very slow to process, will have a quadratically number of parameters with respect to its number of rows and columns and will fill a 16GB RAM very fast (in practice, for less than 10000 neurons per layer given all the other necessary information which have to be stored in the computer memory). To avoid this situation, we have implemented in Cython the computations necessary for the batch weight updates. In this way, we compute in a much faster manner than in pure Python the gradient updates just for the existing connections. For this step, the sparse weight matrices are stored and used in the Coordinate list (COO) format.

\textit{Backpropagation phase - weights update.} For this, the sparse weights matrices are used in the CSR format. 

\subsection{Implementation of weight pruning-regrowing cycle}
\label{sec:7}
In this section, we introduce the implementation of weight pruning-regrowing cycle for Eq. (\ref{Eq:pruning}) and Eq. (\ref{Eq:regrowing}.) The key aspect of the SET method that sets it apart from the conventional DNN training is the evolutionary scheme which modifies the connectivity of the layers at the end of every epoch. As the weight evolution routine is executed quite often, the routine needs to be implemented in an efficient manner to ensure that the SET-MLP training can be done as fast as possible. Furthermore, as the layer connections are extremely sparse in the SET scheme, the implementations should ensure that the sparsity level is maintained. Actually, it shall exploit the sparsity while removing and adding new weights. Two implementations of the weight evolution scheme were coded in native Python using Numpy sparse matrix routines. 

\subsubsection{Implementation I} 
The first implementation is readable and intuitive, but does not exploit the full capabilities of the Numpy library in its various operations. In this implementation, the sparse weight matrices in the CSR format are converted to three vectors representing the indices of the rows, columns of the non-zero elements along with the element values (either using the COO or LIL format). The values are then compared in a for-loop to the threshold to keep the weights or discard them, as per the user specified $\zeta$ values. To ensure that the total number of non-zeros in the weight matrix remains the same, random connections between neurons need to be created. Again a for-loop is used to create new random connections in an incremental manner and ensure that the total number of non-zeros is equal to the original number of non-zeros. 
Most of the processing time in the code occurs in the for-loops and the while loops and this is confirmed by a code profiling tool in python\footnote{Line\_profiler by Robert Kern, [Available Online] \url{https://github.com/rkern/line\_profiler}}. Furthermore, as we are constantly accessing the weights by the row and column index, this method does not exploit the sparsity of the weight matrix. The code profile of the processing time demonstrated that the removal of weights of the weight matrix takes about 15\% of total time in an epoch and adding new random connections takes about 50\% of the total time during an epoch. The detailed algorithm is given in Algorithm \ref{Alg: 2}.

\begin{algorithm}[ht!]
\DontPrintSemicolon
   \SetKwInOut{Input}{Input}
    \SetKwInOut{Output}{Output}
    \Input{Sparse Weight Matrix ($W$)}
    \Output{ Sparse Weight Matrix with random weights added} 
\textit{\%Removal of small weights};\\
Extract values ($V$), row ($R$) and column indices ($C$) of the non-zeros from $W$;\\
Find maximum negative value ($V_{neg}$) and minimum positive value ($V_{pos}$);\\
Initialize $N = 0$;\\
\For{\textit{i in R}}{
\For {\textit{c in C}}{
\If{\textit{$V_{neg} < W_{r,c} < V_{pos}$}}{
 $W_{r,c} = 0$\\
 $N = N + 1$
 }
}
}
{\textit{\%Addition of random weights }}\\
\While{$N > 0$}{
{Choose $i$ randomly from 1 to rows of $W$}\\
{Choose $j$ randomly from 1 to columns of $W$}\\
\If{$W_{i,j} == 0$}{
{Add a random value to $W_{i,j}$}\\
{$N = N - 1$}
}
}
\caption{Weight pruning-regrowing cycle - Implementation I}\label{Alg: 2}
\end{algorithm}


\begin{algorithm}[ht!]
\DontPrintSemicolon
   \SetKwInOut{Input}{Input}
    \SetKwInOut{Output}{Output}
    \Input{Sparse Weight Matrix ($W$)}
    \Output{ Sparse Weight Matrix with random weights added} 
{\textit{\%Removal of small weights}}\;\\
  {Extract values ($V$), row ($R$) and column indices ($C$) of the non-zeros from $W$}\;\\
  {Find the maximum negative value ($V_{neg}$) and the minimum positive value ($V_{pos}$)};\\
  {Find the index $i_{zero}$ of the values ($V$) which are bigger than $V_{neg}$ and smaller than $V_{pos}$};\\
  {Delete the indices of $V$, $R$ and $C$ corresponding to the index $i_{zero}$};\\
  {N = length($i_{zero}$)};\\
{\textit{\%Addition of random weights}};\\
  {Create a list of arrays ($L_{old}$) with the remaining  elements after removing: $L_{old} = [R, C]$};\\
\While{$N > 0$}{
 {I = array of N randomly chosen from 1 to rows of $W$};\\
 {J = array of N randomly chosen from 1 to columns of $W$};\\
 {Create list ($L_{new}$) of arrays with k elements: $L_{new} = [I, J]$};\\
 {Remove duplicate elements from $L{new}$};\\
 {Remove elements from $L_{new}$ in common with $L_{old}$};\\
 {N=N-length($L_{new}$)};\\
 {$L_{old}$ = append ($L_{old}$, $L_{new}$)};\\
 {Clear $L_{new}$};\\
}
 {Append $N$ random values to $V$: $V_{new} = append(V, rand(size=(1,N)))$};\\
 {Unzip 1st and 2nd elements of $L_{old}$: $R_{new} = L_{new}(:,1), C_{new} = L_{new}(:,2)$};\\
 {Use COO format to update the $W$: $W = COO(V_{new},( R_{new}, C_{new}))$ };\\
\caption{Weight pruning-regrowing cycle - Implementation II}\label{Alg: 3}
\end{algorithm}

\subsubsection{Implementation II}
In order to make full use of advantages of different sparse matrix formats, we also propose Fast Weights Evolution (FWE). In FWE, the sparse weight matrices in the CSR format are also converted to three vectors representing the indices of the rows, columns of the non-zero elements along with the element values using the COO format. The value vector is compared a single time with the minimum and maximum threshold values using the vectorized operations in Numpy. This enables the identification of the indices of small weights for fast deletion of the weights. Next, the remaining row and column indices are stored together into an array and a list of all the arrays of the non-zero elements is created. This is used directly to determine the random row and column indices of the additional weights to ensure that the number of connections between the neurons is constant. As the weights are sparse, the size of the list is much smaller than the full size of the weight matrix and performing all the computations with the list will be faster. The detailed algorithm is given in Algorithm \ref{Alg: 3}. The comparison of the running time of these two implementations is given in Table \ref{Imcomparison}, which shows that Implementation II is more efficient than Implementation I. The computational complexity (Big O Notation) is the same for both implementations. The difference is given by running python code without optimised C++ routines (Algorithm \ref{Alg: 2}) and with optimised C++ routines (Algorithm \ref{Alg: 3}).

\begin{table}
\small
\caption{\label{Imcomparison}Mean running time of evolution Implementation I and Implementation II.}
\begin{center}
\setlength{\tabcolsep}{1mm}{
\begin{tabular}{ |c |r |r |}
\hline
\textbf{Matrix Size}  & \textbf{Implementation I (s)} & \textbf{Implementation II (s)} \\
\hline
500*500 & 0.58 & 0.14 \\
\hline
2000*2000 & 2.56 & 0.71 \\
\hline
8000*8000 & 11.13 & 2.08 \\
\hline
15000*15000 & 24.14 & 3.75 \\
\hline
\end{tabular}}
\end{center}
\end{table}

\section{Experimental evaluation}
\label{sec:8}
For a good understanding of SET-MLP performance, we compare it against another sparse MLP model (implemented by us in the same manner) in which the bipartite layers are initialized with an Erd\H{o}s-R\'{e}nyi topology, but does not evolve over time and has a fixed sparsity pattern, dubbed MLP$_{FixProb}$ as in~\cite{mocanu2018scalable}. Note that it is impossible to report also the accuracy for FC-MLPs as they can not run on a typical laptop due to their very high memory and computational requirements. Moreover, even if it would be possible to run FC-MLP, this comparison is outside the scope of this paper and it would be redundant as it has been shown in \cite{evci2020rigging,liu2019improving,mocanu2018scalable,mocanu2017network,zhu2019multi,bourgin2019cognitive} that SET-MLP typically outperforms its fully-connected counterparts.
\subsection{Datasets}
We evaluate and discuss the performance of our efficient SET-MLP implementation on four publicly available microarray datasets, as detailed in Table \ref{dataset}. It is worth highlighting that both, the training and testing sets, are unbalanced for all datasets. We choose $2/3$ of the data as training data and $1/3$ of the data as testing data. Note that we do not set validation data, as the sample sizes of these datasets are extremely small.
\begin{table}
\caption{\label{dataset}Microarray datasets used.}
\scriptsize
\begin{center}
\setlength{\tabcolsep}{1mm}{
\begin{tabular}{ l l l l l}
\hline
\textbf{Dataset} & \textbf{No. of} & \textbf{No. of} & \textbf{No. of} & \textbf{Data} \\
 & \textbf{Samples} & \textbf{Features} & \textbf{Classes} & \textbf{Size} \\
\hline
Leukemia \cite{haferlach2010clinical} & 2096& 54,675& 18& 1.93 GB \\
CLL-SUB-111 \cite{haslinger2004microarray} & 111& 11,340& 3& 5.9 MB \\
SMK-CAN-187 \cite{zhao2010advancing} & 187 & 19,993 & 2 & 11.9 MB\\
GLI-85 \cite{freije2004gene}  & 85 & 22,283 & 2 & 8.7 MB\\
\hline
\end{tabular}}

\end{center}
\vspace{-2em}
 \end{table} 

\paragraph{Leukemia} The Leukemia dataset is obtained from the NCBI GEO repository with the accession number GSE13159. It contains 2096 samples with 54,675 features each. The samples are divided into 18 classes. Among these 2096 samples, 1397 samples are selected as training data and 699 as testing data. Table \ref{2} shows the number of test samples in each class.
\begin{table}
\caption{\label{2}Leukemia class labels and their corresponding number of test samples.}
\scriptsize
\begin{center}
\setlength{\tabcolsep}{1mm}{
\begin{tabular}{ l l l}
\hline
\textbf{Leukemia} & \textbf{Class} & \textbf{No. of} \\ 
& \textbf{Label} & \textbf{Samples}\\ 
\hline
Mature B-ALL with t(8;14) & 1 & 4\\
Pro-B-ALL with t(11q23)/MLL & 2 &23\\
C-ALL/Pre-B-ALL with t(9;22) & 3 & 41\\
T-ALL & 4 & 58\\
ALL with t(12;21) & 5 & 19 \\
ALL with t(1;19) & 6 & 12 \\
ALL with hyperdiploid karyotype & 7 & 14 \\
C-ALL/Pre-B-ALL without t(9;22) & 8 & 79\\
AML with t(8;21) & 9 & 14\\
AML with t(15;17) & 10 & 12\\
AML with inv(16)/t(16;16) & 11 & 9\\
AML with t(11q23)/MLL & 12 & 13\\
AML with normal karyotype + other abnormalities & 13 & 115\\
AML compllex aberrant Karyotype & 14& 18 \\
CLL & 15 & 149\\
CML & 16 & 25\\
MDS & 17 & 68\\
Non-leukemia and helthy bone marrow & 18 & 26\\
\hline
\end{tabular}}
\end{center}
\vspace{-2em}
\end{table}
\paragraph{CLL-SUB-111} The CLL-SUB-111 dataset is an unbalanced dataset contains gene expressions from high-density oligonucleotide arrays consist of both genetically and clinically distinct subgroups of B-cell chronic lymphocytic leukemia (B-CLL). It has 11,340 features and 111 samples, out of which 74 samples are selected as the training set and 37 as the testing set.
\paragraph{SMK-CAN-187} The SMK-CAN-187 dataset is a RNA dataset obtained from the normal bronchial epithelium of smokers with and without lung cancer. It has 19,993 features and 187 samples. Out of these 187 samples, 124 samples are chosen as training data and 63 as testing data.
\paragraph{GLI-85} The GLI-85 dataaset is the Affymetrix HG U133 oligonucleotide arrays on 85 diffuse infiltrating gliomas of all histologic types. It has 22,283 features and 85 samples. Out of these 85 samples, 56 samples are training data and 29 are testing data.

\begin{figure*}[ht]
 \centering
 \begin{subfigure}[b]{0.4\textwidth}
\includegraphics[width=1\textwidth]{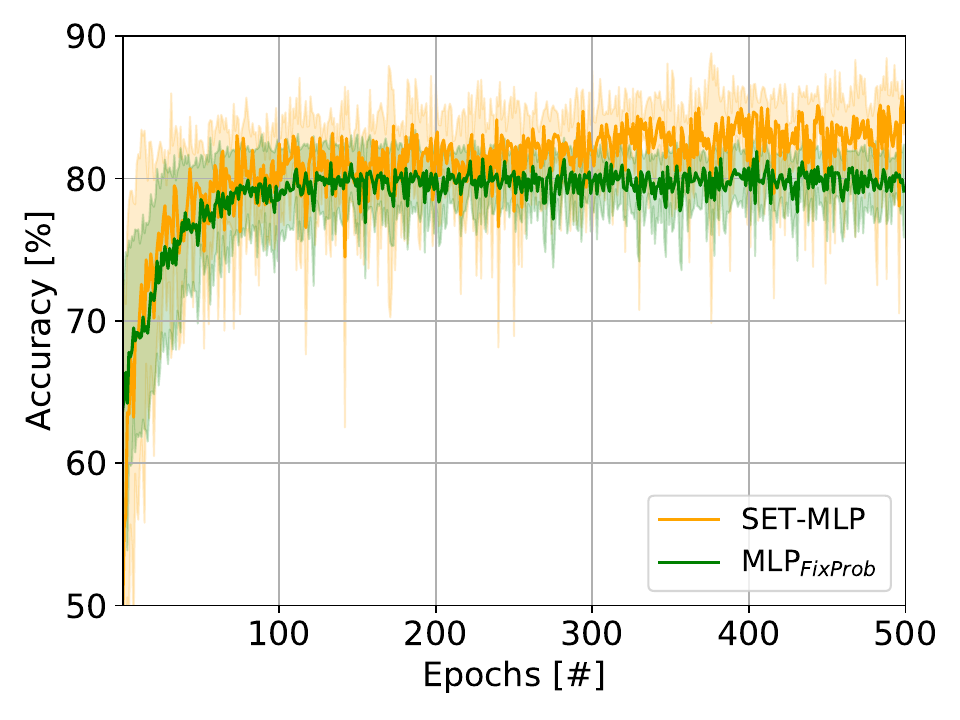}
\caption{\label{Leukemia_accuracy}Leukemia}
\end{subfigure}
\hspace*{0.5cm} 
 \begin{subfigure}[b]{0.4\textwidth}
\includegraphics[width=1\textwidth]{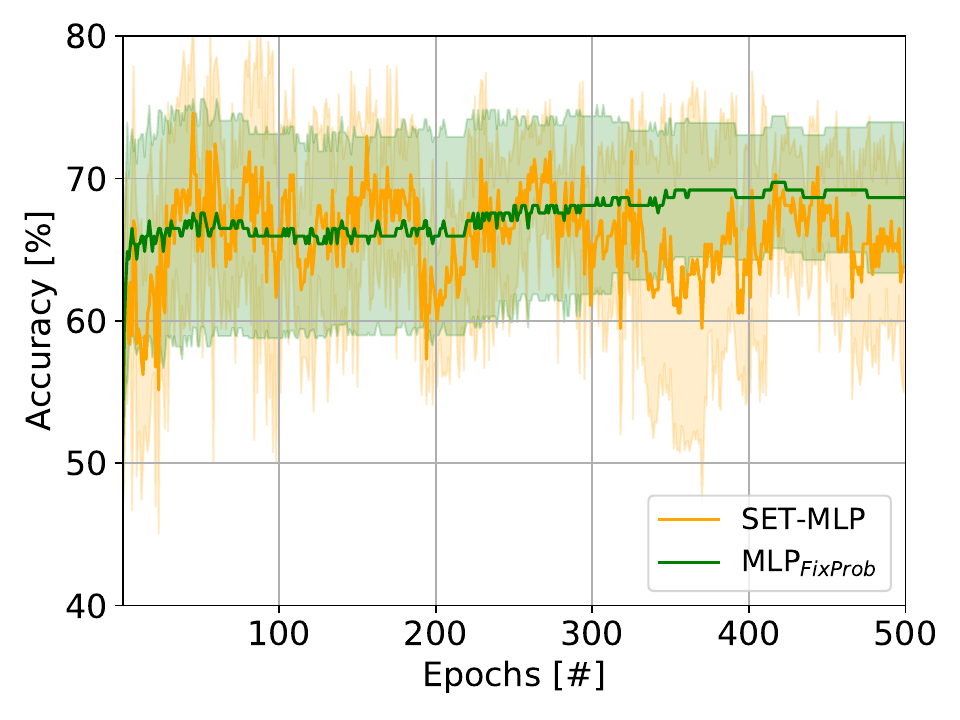}
\caption{\label{CLL_accuracy}CLL-SUB-111}
\end{subfigure}
 \begin{subfigure}[b]{0.4\textwidth}
\includegraphics[width=1\textwidth]{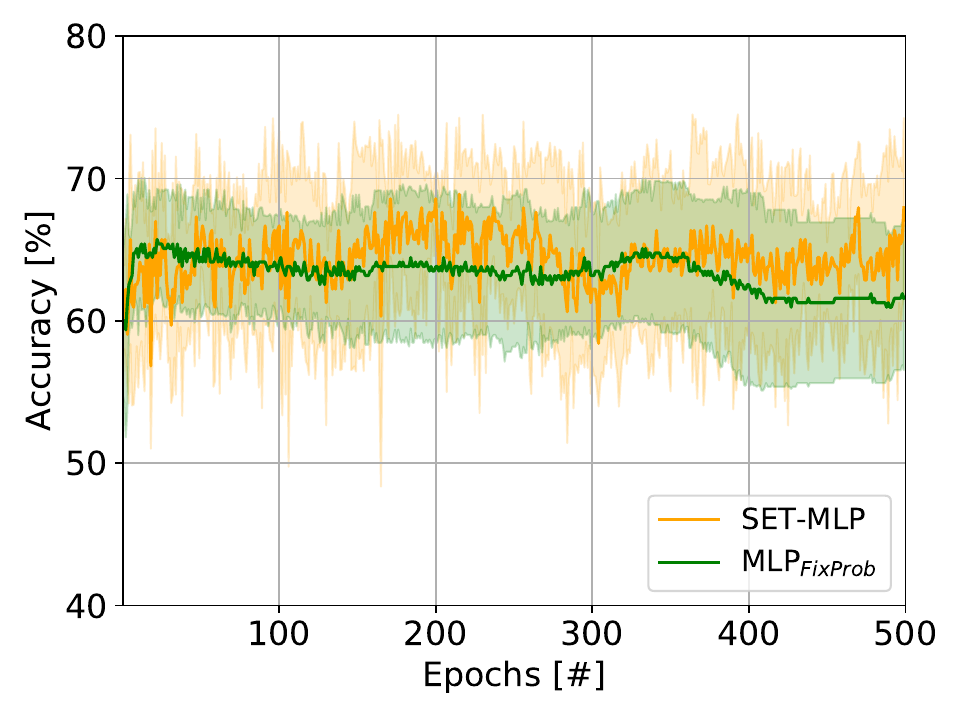}
\caption{\label{SMK_accuracy}SMK-CAN-187}
\end{subfigure}
\hspace*{0.5cm} 
 \begin{subfigure}[b]{0.4\textwidth}
\includegraphics[width=1\textwidth]{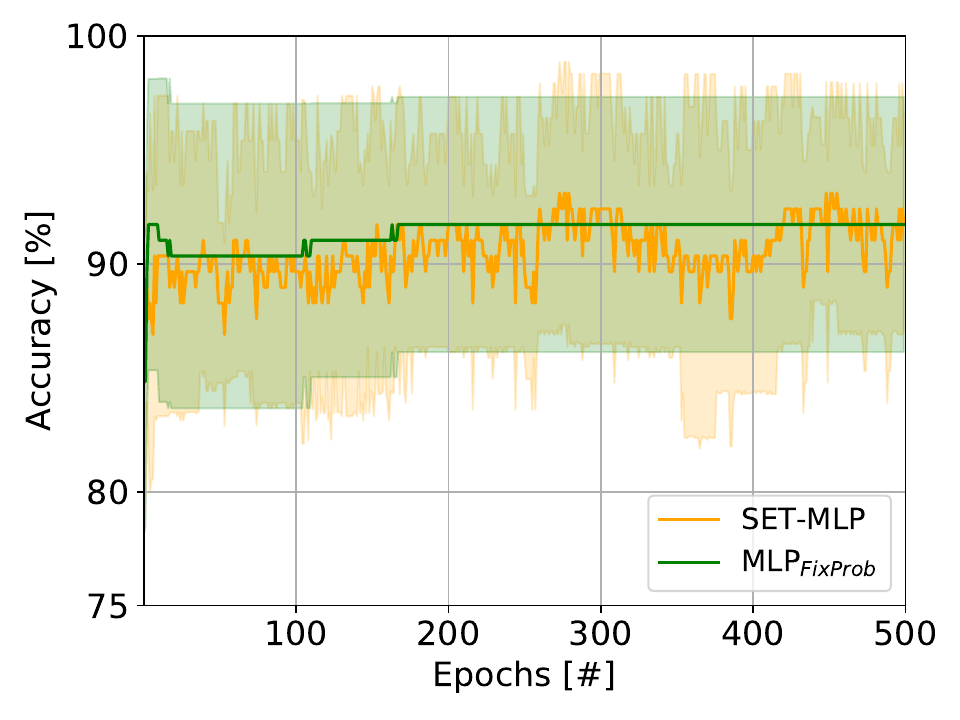}
\caption{\label{Gli_accuracy}GLI-85}
\end{subfigure}
\caption{\label{learning_curve}Test accuracy of SET-MLP and MLP$_{FixProb}$ on CLL-SUB-111, SMK-CAN-187 and GLI-85. All the test accuracy at each epoch is averaged from 5 trials.} 
\end{figure*}
\subsection{Evaluation metrics}
\label{sec:9}
To evaluate the performance of the proposed method, we have used the accuracy metric and the confusion matrix to get detailed visual information. The confusion matrix ($\mathbf{M}$) contains information per class about both, model predictions and ground truth. These enable people to understand and diagnose the models better. The confusion matrix template and related performance measures for two-class classification problems are given in Table \ref{confusion}. In terms of multi-class
classification, assuming that the number of classes is $c$, the performance measures of the $i^{th}$ class are given by the following equations:
\begin{equation}
\begin{aligned}
Recall_i & =\frac{M_{ii}}{\sum_{j=1}^{c}M_{ji}} \\
Precision_i & =\frac{M_{ii}}{\sum_{j=1}^{c}M_{ij}} \\
Accuracy & =\frac{\sum_{i=1}^cM_{ii}}{\sum_{i=1}^c\sum_{j=1}^cM_{ij}}
\end{aligned}
\end{equation}
\begin{table*}[ht!]
\centering
\caption{Confusion matrix for two class classification. $Spe$, $Rec$, $Pre$ and $Acc$ are acronyms for Specificity, Recall, Precision and Accuracy, respectively.}
\label{confusion}
\begin{center}
\begin{tabularx}{0.65\textwidth}{lllll}
\hline
                         &  & \textbf{Target Class} &              &  \\ \cline{3-4}
                         &  & \textbf{Neg}          & \textbf{Pos} &  \\ \hline
Output Class             &  &                       &              &  \\
Classified as \textbf{Neg} & & \(tn\)                & \(fn\)           &   \(npv = \frac{tn}{tn+fn} \)\\
Classified as \textbf{Pos} & & \(fp\)                & \(tp\)           &   \(Pre = \frac{tp}{tp+fp} \)\\
                           & & \(Spe = \frac{tn}{tn+fp} \)   & \(Rec  = \frac{tp}{tp+fn} \) & \(Acc = \frac{tp+tn}{tp+fp+fn+tn} \) \\ 

\hline
\end{tabularx}
\end{center}
\end{table*}

The rows of the confusion matrix represent the predicted classes and the columns correspond to the true classes. The diagonal cells represent the numbers of samples that are correctly classified. The off-diagonal cells are the incorrectly classified number of samples. The row at the bottom of the confusion matrix gives the proportion of all examples belonging to each class that is correctly (green) and incorrectly (red) classified.
The column on the far right of the confusion matrix represents the proportion of all the samples predicted to belong to each class that are correctly (green) and incorrectly (red) classified. 

\subsection{Experimental setup}
 \begin{figure*}[ht]
 \centering
  \begin{subfigure}[b]{0.3\textwidth}
\includegraphics[width=1.1\textwidth]{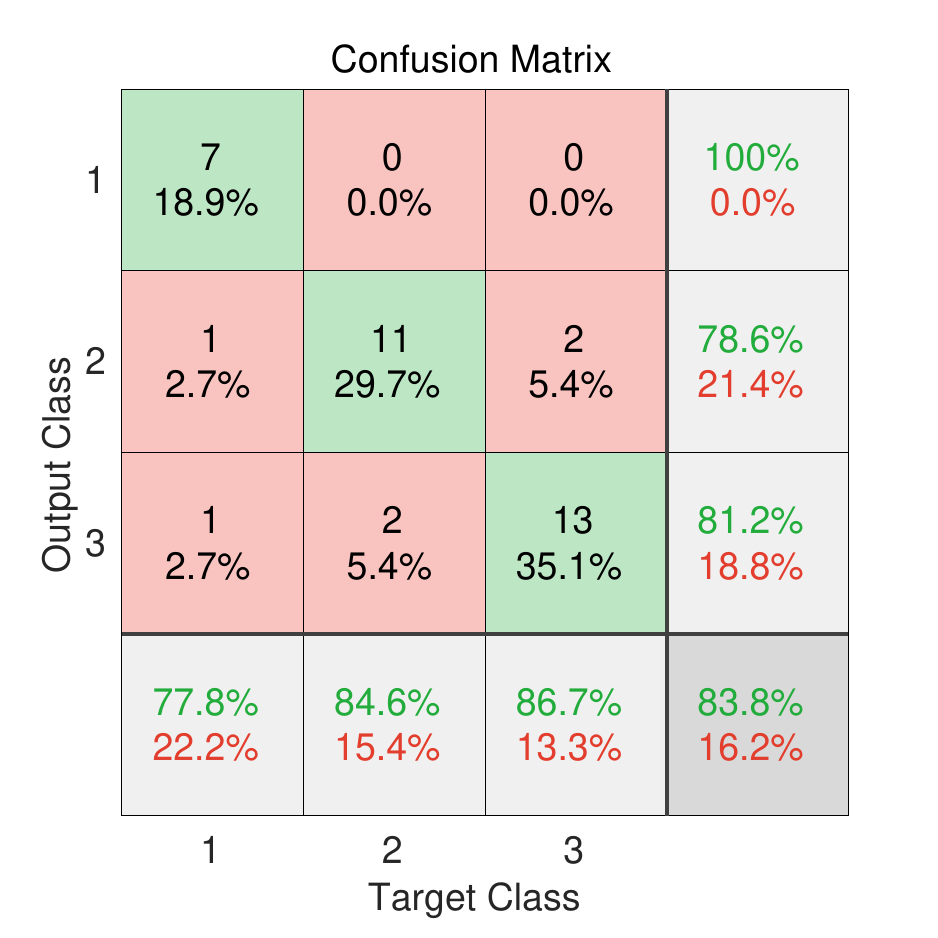}
\caption{\label{cll_confusion}CLL-SUB-111 dataset}
\end{subfigure}
 \begin{subfigure}[b]{0.3\textwidth}
 \includegraphics[width=1.1\textwidth]{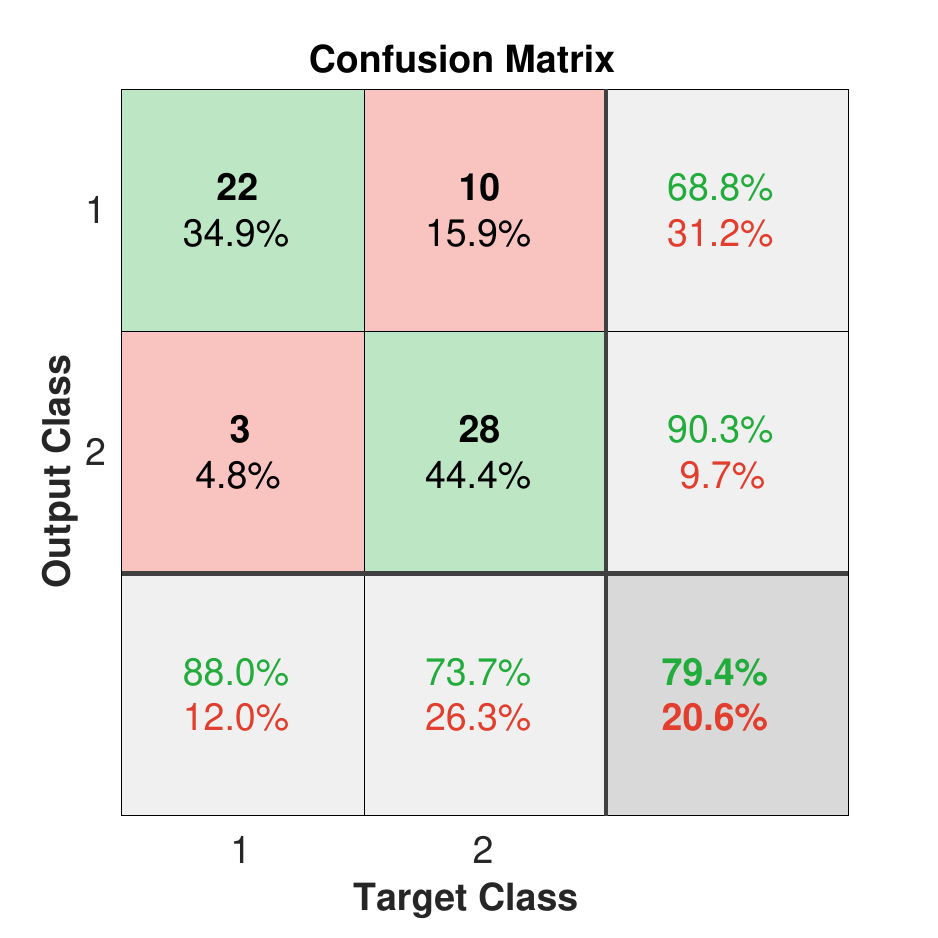}
\caption{\label{sml_confusion}SMK-CAN-187 dataset}
\end{subfigure}
 \begin{subfigure}[b]{0.3\textwidth}
\includegraphics[width=1.1\textwidth]{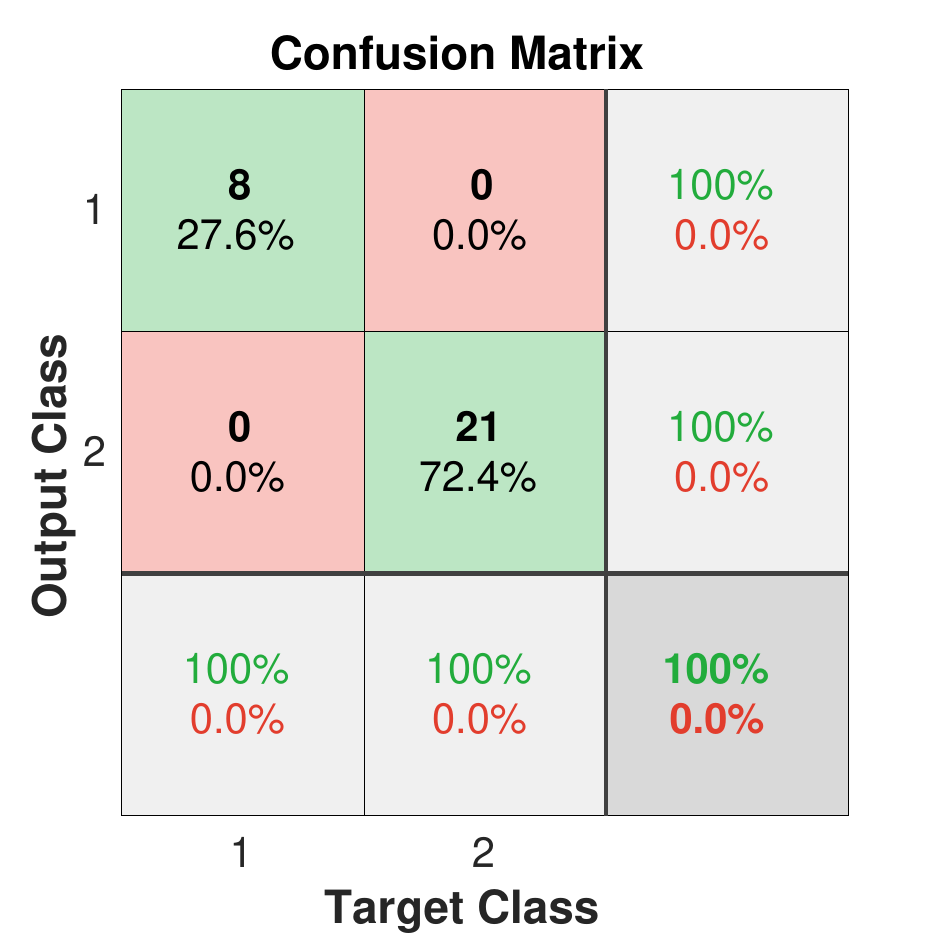}
\caption{\label{GLI_85_confu}GLI-85 dataset}
\end{subfigure}
\caption{\label{rest_confu}Confusion matrix of the best run with SET-MLP on CLL-SUB-111, SMK-CAN-187 and GLI-85.}
\end{figure*}
For both models, SET-MLP and MLP$_{FixProb}$, the hyperparameters are the same to guarantee the fairness of comparison. The number of hidden layers for Leukemia, CLL-SUB-111 and SNK-CAN-187 is two but one for GLI-85, as overfitting occurs for GLI-85 with two hidden layers. The optimization method used in this paper is Stochastic Gradient Descent (SGD) with momentum. The numbers of neurons of each layer are given by Table \ref{numberofneurons}. Note that, for Leukemia dataset, the number of hidden neurons in each layer was set to 27,500, a value which is way above the usual number of neurons in fully-connected MLP models. For all datasets, we get the mean accuracy by averaging the best test accuracy from 5 trials. Since the best accuracy is obtained at different epochs, there are some differences between accuracy in the figures and the mean accuracy reported in the text, especially for the CLL-SUB-111 and the SMK-CAN-187 datasets.
\begin{table}[ht]
\scriptsize
\caption{\label{numberofneurons}Number of neurons of SET-MLP on all datasets.}
\begin{center}
\begin{tabular}{c| r| r| r| r}
\hline
\textbf{Dataset}  & \textbf{Input} & \textbf{$1^{st}$ hidden}& \textbf{$2^{nd}$ hidden}& \textbf{Output}\\
\hline
Leukemia & 54,675 & 27,500 & 27,500 & 18\\
\hline
CLL-SUB-111  & 11,340  & 9,000 & 9,000 & 3\\
\hline
SNK-CAN-187 & 19,993 & 16,000 & 16,000 & 2\\
\hline
GLI-85 & 22,283 & 20,000 & none & 2\\
\hline
\end{tabular}
\end{center}
\end{table}

To demonstrate our algorithm's ability to significantly reduce the parameter count, we set the sparsity hyperparameter $\epsilon = 10$ guaranteeing an extremely sparse network for all datasets. The corresponding sparsity for Leukemia, CLL-SUB-111, SNK-CAN-187, and GLI-85 is 99.93\%, 99.78\%, 99.88\%, and 99.90\%, respectively. The corresponding sparsity and parameter numbers are illustrated in Table \ref{sparsity_paramters}. The rewiring rate $\zeta$ is set to 0.3. We train all models for 500 epochs by Momentum SGD with a momentum of 0.9 and a weight decay of 0.0002. We choose the remaining hyperparameters based on a small random search. For Leukemia, we use a learning rate of 0.005 and a batch size of 5; For CLL-SUB-111, we choose a learning rate of 0.01 and a batch size of 5; For SNK-CAN-187, the learning rate is set as 0.005 and the batch size is set as 5; For GLI-85, the learning rate is set as 0.005 and the batch size is set as 1.

All the experiments performed are executed on a typical laptop using \textit{a single thread of the CPU}. The laptop configuration is as follows:
\begin{itemize}
\item Hardware configuration: CPU Intel Core i7-4700MQ, 2.40 GHz $\times$ 8, RAM 16 GB, Hard disk 500 GB.
\item Software used: Ubuntu 16.04, Python 3.5.2, Numpy 1.15.2, SciPy 1.1.0, and Cython 0.27.3.
\end{itemize}
\subsection{Experimental results}

\begin{table*}[ht]
\caption{\label{Testaccuracy}Test accuracy of SET-MLP and MLP$_{FixProb}$ on the four datasets. Every number is averaged from 5 trials. The sparsity levels for Leukemia, CLL-SUB-111, SNK-CAN-187 and GLI-85 are 99.93\%, 99.78\%, 99.88\% and 99.90\%, respectively. }
\begin{center}
\begin{tabular}{l| c| c| c| c}
\hline
\textbf{Methods}  & \textbf{Leukemia} & \textbf{CLL-SUB-111}& \textbf{SNK-CAN-187} & \textbf{GLI-85}\\
\hline
SET-MLP & \textbf{$87.60 \pm 0.06$} & $81.62 \pm 0.05$ & $75.24\pm 0.04$ & $94.48 \pm 0.05$\\
\hline
MLP$_{FixProb}$  & $82.74 \pm 0.05$ & $71.35 \pm 0.04$ & $68.57 \pm 0.04$ & $92.41 \pm 0.06$ \\
\hline
\end{tabular}
\end{center}
\end{table*}

Table \ref{Testaccuracy} summarizes the performance of SET-MLP and MLP$_{FixProb}$ on all four datasets trained with extremely high sparsity levels. We can observe that SET-MLP consistently outperform MLP$_{FixProb}$ on all datasets, which means that the adaptive sparse connectivity associated with SET-MLP helps to find better sparse structures. This behavior suggests that the SET algorithm indeed solves successfully the combinatorial optimization problem. From the perspective of continuous optimization, the optimizer used in this experiment, momentum SGD, is used for optimizing model weights. Both, SET and momentum SGD, are crucial to the superior performance of sparse training\cite{mostafa2019parameter}.

For a better understanding of the learning process of our method, we show the learning curves of SET-MLP and MLP$_{FixProb}$ for all datasets in Figure \ref{learning_curve}. It is shown that SET-MLP can reach a higher accuracy than MLP$_{FixProb}$ as the training epoch increases. More interestingly, the learning curves of SET-MLP oscillate more frequently than the fixed sparse networks during the training process. This phenomenon makes sense since the weights rewiring ( pruning and regrowing) cycle within adaptive sparse connectivity is triggered after each training epoch, changing 30\% connections of the network. Furthermore, it is noteworthy that, on SMK-CAN-187 dataset, MLP$_{FixProb}$ seems to suffer from overfitting after around 350 epochs.

\begin{figure}[ht!]
\includegraphics[width=0.5\textwidth]{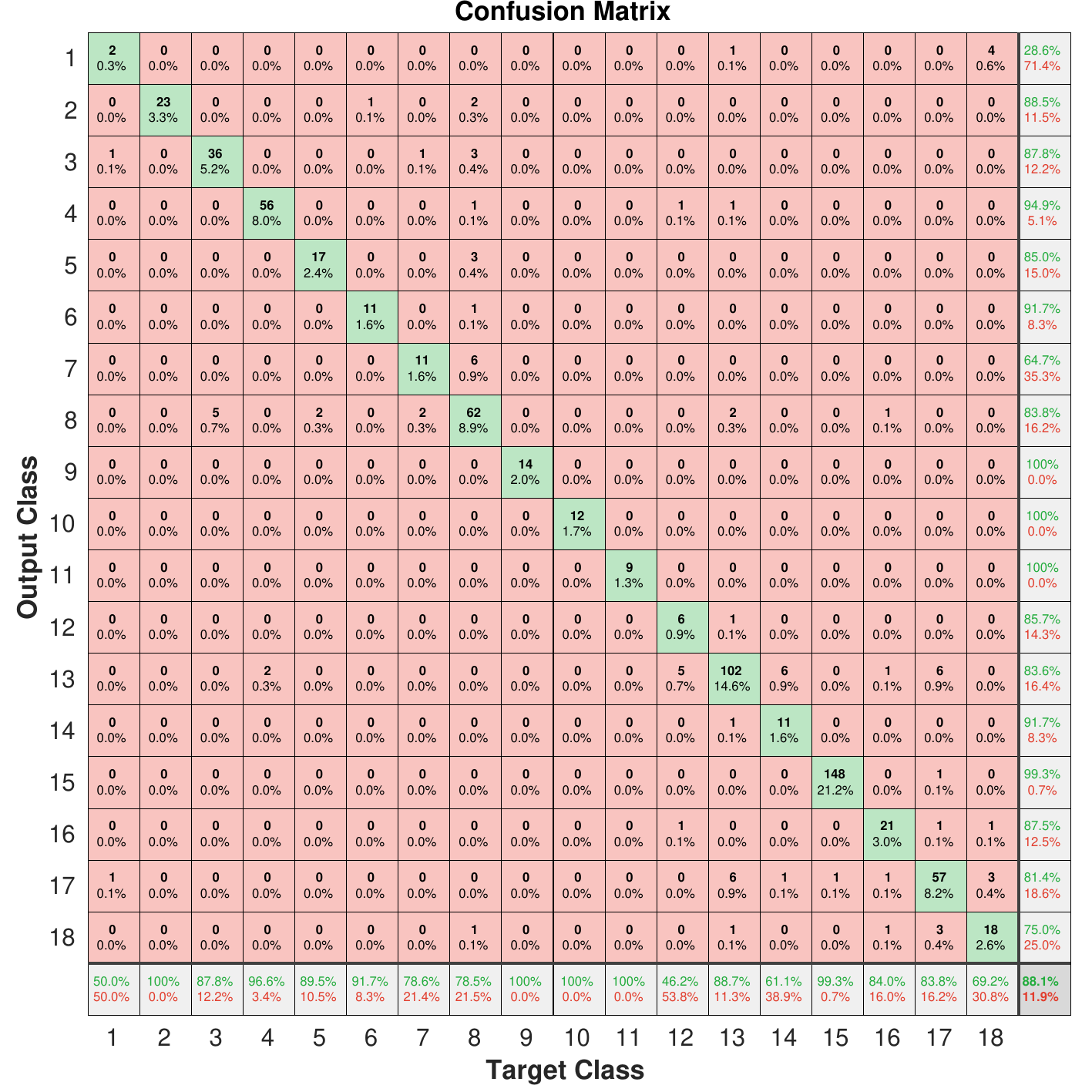}
\caption{\label{leukemia_confu}Confusion matrix of the best run with SET-MLP on the Leukemia dataset.}
\end{figure}

To provide deeper insights into the classification results of our method, we illustrate the confusion matrices of the best run of SET-MLP on Leukemia dataset in Figure \ref{leukemia_confu} and the rest of the datasets in Figure \ref{rest_confu}. We can see that the test accuracy of the best run of Leukemia is 88.10\%. Besides this, we can observe that SET-MLP has a perfect recall for class 1 (100.0\%) of CLL-SUB-111, even though there are extremely unfavorable conditions, i.e. very few training samples. As shown in Figure \ref{sml_confusion}, SET-MLP performs better for class 2 than class 1 on SNK-CAN-187. It is noteworthy that, for GLI-85, the best accuracy out of five runs of SET-MLP is 100\%, which means on Gli-85 whose available data are extremely insufficient (85), SET-MLP can still model the dataset perfectly.

To further evaluate the effectiveness of our proposed method, we compare SET-MLP with the state-of-the-art conventional two-phase techniques on these datasets in terms of classification performance. To the best of our knowledge, the state-of-the-art performance for Leukemia is 81.11\% reported in the literature \cite{kumar2015classification}. Therein, an ensemble classifier is proposed to deal with microarray data by connecting several feature selection algorithms with MapReduce based proximal support vector machine (mrPSVM). SET-MLP is able to achieve a higher accuracy of $87.60 \pm 0.06$ with exactly the same training and testing data splitting. Among the feature selection based methods to CLL-SUB-111, an accuracy of 78.38\% is obtained by using Incremental Wrapper-based Attribute Selection(IWSS) \cite{bermejo2012fast}. The state-of-the-art accuracy on SMK-CAN-187 is (74.87$\pm$2.32\%) reported in \cite{wang2016supervised}, in which feature selection was performed by preserving class correlation. Reported in \cite{taheri2014hybrid}, an ensemble including three filter methods with a meta-heuristic algorithm is used to achieve an accuracy of 94\%. We can observe that our method can outperform these traditional two-phase techniques via one efficient end-to-end model. It is noteworthy that, although CLL-SUB-111, SMK-CAN-187 and GLI-85 seriously suffer from an extremely small number of samples, we are still able to obtain good performance with efficient sparse training. 

\subsection{Results analysis}
\begin{figure}[ht]
\centerline{\includegraphics[width=0.45\textwidth]{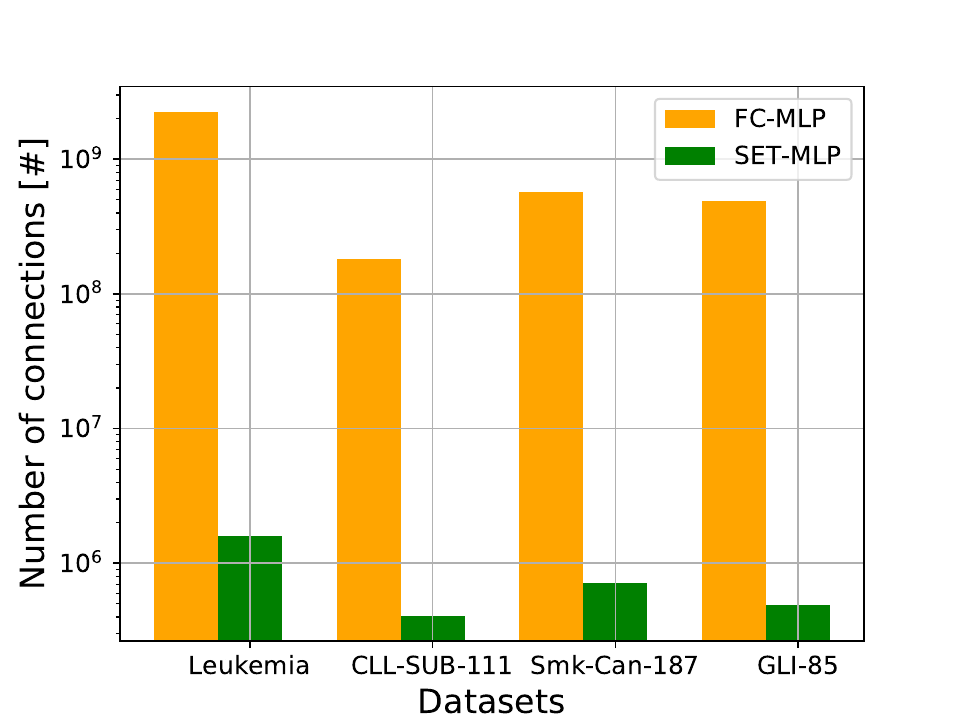}}

\caption{\label{connectionnumber}The number of connections for the SET-MLP models with two hidden layers used on the Leukemia, CLL-SUB-111, and SMK-CAN-187 datasets and with one hidden layer used on the GLI-85 dataset, plotted against their FC-MLP counterparts.}
\end{figure}

To understand better the connections reduction made by the SET procedure in a SET-MLP model in comparison with a fully-connected MLP (FC-MLP) which has the same amount of neurons, Figure \ref{connectionnumber} and Table \ref{sparsity_paramters} provide the number of connections for the SET-MLP models discussed above and their FC-MLP counterparts on all four datasets. It is clear that SET has dramatically reduced the connection numbers in MLPs. For instance, a traditional FC-MLP on the Leukemia dataset would have 2,260,307,500 connections, while SET-MLP has just 1,582,376 connections. This quadratic reduction in the number of connections is the key factor in guaranteeing that SET-MLP can run fine on a standard laptop for datasets with tens (up to few hundreds) of thousands of input features. 

\begin{table}[ht]
\small
\caption{\label{sparsity_paramters}Number of connections and sparsity levels for SET-MLP and FC-MLP on all datasets.}
\begin{center}
\begin{tabular}{ r |r| r |r }
\hline
\textbf{Dataset}  & \multicolumn{2}{r|}{\textbf{Number of connections (\#)}}& \textbf{Sparsity}\\
  & \textbf{FC-MLP} & \textbf{SET-MLP}& \textbf{level} \\
\hline
Leukemia & 2,260,307,500 & 1,582,376 & 99.93\%\\
\hline
CLL-SUB-111  & 183,087,000  & 409,033 & 99.78\%\\
\hline
SMK-CAN-187 & 575,920,000 & 711,305 & 99.88\%\\ 
\hline
GLI-85 & 490,270,000 & 486,350 & 99.90\%\\ 
\hline
\end{tabular}
\end{center}
\end{table}

\begin{table}[t]
\small
\caption{\label{5}Averaged running time in seconds (s) per epoch for SET-MLP.}
\begin{center}
\begin{tabular}{ |l| r| r| r|}
\hline
\textbf{Dataset}  & \textbf{Training time (s)} & \textbf{Testing time (s)} \\
&  \textbf{(per epoch)} & \textbf{(per epoch)}\\
\hline
Leukemia & 61.31 & 2.36\\ 
\hline
CLL-SUB-111  & 6.65  & 0.06\\
\hline
SMK-CAN-187 & 27.17 & 0.18\\
\hline
GLI-85 & 32.87 & 0.05\\
\hline
\end{tabular}
\end{center}
\end{table}

For a better understanding of SET computational requirements, Table \ref{5} shows the average training and testing time per epoch of the SET-MLPs used on the datasets. We can observe, as expected, that as the number of features and samples increases the training time is also increasing. Still, it is worth to highlight that, although the average training time of Leukemia is relatively long (61.31s), it fulfills an almost impossible mission, that is, running such a large model on a commodity laptop.   
\begin{figure*}[ht!]
\centerline{\includegraphics[width=0.7\textwidth]{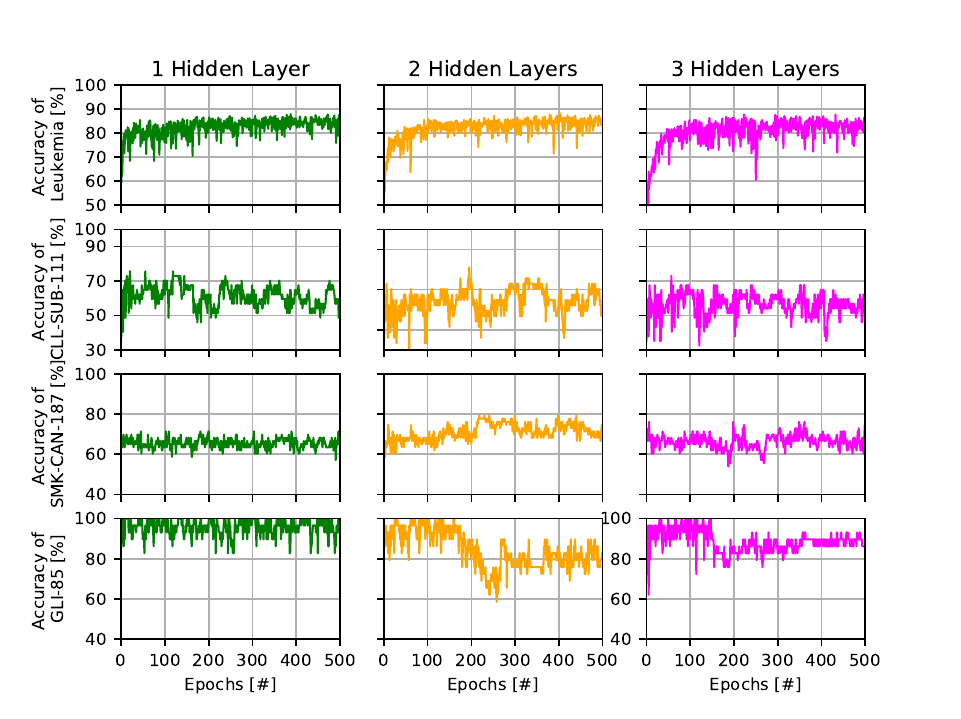}}
\caption{\label{comparisonwithhl}Experiments with SET-MLPs on all four datasets to understand the effect of the number of hidden layers($n^h$). For each dataset, three cases for the number of hidden layers are considered, i.e. $n^h = \{1, 2, 3\}$. Each row represents the test classification accuracy of SET-MLPs with one, two, or three hidden layers on the same dataset. Every model from each row has been trained with the same hyperparameters as in the paper, except for the number of hidden layers. }  
\end{figure*}

\begin{figure}[ht!]
 \centering
 \begin{subfigure}[b]{0.4\textwidth}
\includegraphics[width=1\textwidth]{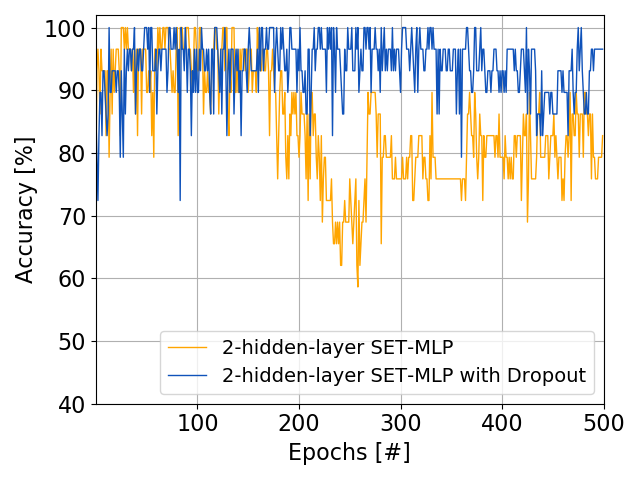}
\caption{\label{dropout_gli85}}
\end{subfigure}
 \begin{subfigure}[b]{0.4\textwidth}
\includegraphics[width=1\textwidth]{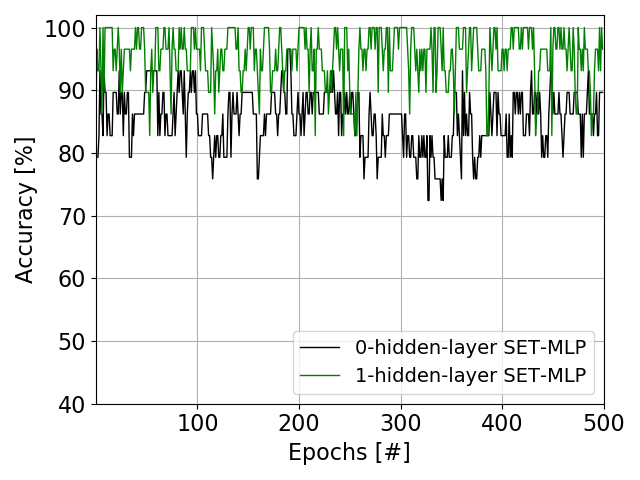}
\caption{\label{LR}}
\end{subfigure}
\caption{\label{LR_Dropout}Test accuracy of SET-MLP and MLP$_{FixProb}$ on GLI-85. All the test accuracy at each epoch is averaged from 5 trials.} 
\end{figure}
\subsection{Extreme SET-MLP models on Leukemia}
\label{sec:discextreme}
While in the previous section, we have analyzed the qualitative performance of our proposed approach, in this section, we briefly discuss two extreme SET-MLP models on the largest dataset used in this paper, i.e. Leukemia. The goal is to assess how fast SET-MLP can achieve a good performance and to see how large a trainable SET-MLP model can be on a typical laptop. For each model, we used a SET-MLP with two hidden layers, and a Softmax layer as output. For the small SET-MLP model, the number of hidden neurons per layer was set to 1,000, while for the large SET-MLP model the number of hidden neurons per layer was set to 500,000. In both cases, we have used a very eager learning rate (0.05) and we trained the models for 5 epochs. On each hidden layer, we applied a dropout rate of 0.4. The other hyperparameters were set as in the previous section for Leukemia and we have used the same training/testing data splitting. 

\begin{table*}[ht!]
\caption{\label{tab:extremecases}Two extreme SET-MLP models on Leukemia against state-of-the-art (mrPSVM with ANOVA~\cite{kumar2015classification} for feature selection). The numbers in brackets for SET-MLP reflect the number of neurons per layer from input to output. The accuracy of SET-MLP is reported as the mean and standard deviation of 5 runs. The density level represents the percentage of the number of existing connections in the SET-MLP model from the total number of connections in its corresponding FC-MLP.}

\begin{center}
\small
\begin{tabular}{ |l| l| l|l| l|}
\hline
\textbf{Model}  & \textbf{Hardware} & \textbf{Density} & \textbf{Total time (s)} & \textbf{Accuracy (\%)}  \\
&  & \textbf{level (\%)} & \textbf{(train + test)}&\\
\hline
Small SET-MLP & 1 CPU thread  &1.04& 65 &  82.88$\pm$1.18\\ 
\hline
Large SET-MLP &1 CPU thread &  0.007 & 4914 & 81.83$\pm$1.11\\
\hline
mrPSVM ~\cite{kumar2015classification}  & Conventional & n/a & 1265 & 81.1\\
\hline
mrPSVM ~\cite{kumar2015classification} & Hadoop cluster & n/a& 291 & 81.1\\
\hline
\end{tabular}
\end{center}
\end{table*}

Table~\ref{tab:extremecases} presents SET-MLP performance in comparison with the best state-of-the-art results of mrPSVM from~\cite{kumar2015classification}. We clarify that the goal of this experiment is not to obtain the best accuracy possible with SET-MLP. Still, the small SET-MLP model, which has in total 56,693 neurons and 581,469 connections, has a total training and testing time of 65 seconds. It is about 20 times faster than mrPSVM which runs on conventional hardware and about 4.5 times faster than mrPSVM which runs in a Hadoop cluster while reaching with 1.7\% better accuracy. At the same time, its small standard deviation shows that the model is very stable. Furthermore, we highlight that the very large SET-MLP model which has in total 1,054,693 neurons with about 19,383,046 connections takes about 16 minutes per training epoch and in 5 epochs reaches a good accuracy, better than state-of-the-art. All of these happen on 1 CPU thread of a typical laptop. We highlight that this is the first time in the literature when a MLP variant with over 1 million neurons is trained on a laptop, while the usual MLP models trained on a laptop can have at maximum few thousand of neurons. In fact, it is hard to quantify, but according to~\cite{Goodfellow:2016:DL:3086952}, the size of the largest neural networks which run currently in the cloud is about 10 to 20 million neurons. Therefore, our results emphasize even more the capabilities of SET-MLPs and open the path for new research directions.

\subsection{Sensitivity analysis of the number of hidden layers}
\label{app:b}

Previously, we have discussed the performance of the SET-MLP models with two hidden layers on the Leukemia, CLL-SUB-111, SMK-CAN-187 datasets and with one hidden layer on the GLI-85 datasets. We now explain our choices on the number of hidden layers by presenting the performance of SET-MLP models with one, two, and three hidden layers on all datasets comparatively and by discussing the beneficial effect of dropout~\cite{dropoutpaper} on SET-MLP. The number of neurons per hidden layer and the other hyperparameters are set to be the same with the previous models. Figure \ref{comparisonwithhl} summarizes these experiments. From the first row, it can be inferred that SET-MLP with two hidden layers reaches the highest peak accuracy (88.12\%) and has relatively the most robust performance on the Leukemia dataset. Similarly, SET-MLP with two hidden layers reaches outstanding accuracy (81.11\%) on the CLL-SUB-111 dataset, while the accuracy can not reach 80\% with one or three hidden layers.

As expected, but at the same time having the most interesting results, due to the very small number of samples of GLI-85 (Figure \ref{comparisonwithhl}, third row), 
SET-MLP with one hidden layer avoids overfitting in exchange to quite an oscillating behavior. At the same time, SET-MLP with two or three hidden layers even if they are capable of also reaching perfect accuracy of 100\%, after about 200 epochs, they have a dramatic drop in accuracy to about 80\%. 
We hypothesis that this situation happens due to overfitting as the number of training samples is extremely insufficient. If this is the case, adding dropout regularization to SET-MLP is able to figure out this problem. We applied dropout with 0.5 dropout rate to both hidden layers. The performance is shown in Figure \ref{dropout_gli85}. It is clear that the accuracy of SET-MLP with dropout keeps the same trend as before, without any drop in accuracy after 200 epochs. Moreover, we conduct an extra experiment to test if SET-MLP with no hidden layers can achieve higher accuracy or not. Since the number of input features is much higher than the number of classes, the connectivity is almost dense. As shown in Figure \ref{LR}, it cannot reach 100\% classification performance. This phenomenon highlights the fact that our proposed method can guarantee efficient training while not compromising performance.

\section{Conclusion}
Processing microarray data have been treated in the literature as a difficult task due to their very high number of features but the little number of examples. Besides that, this type of data suffers from imbalance and data shift problems. 

In this paper, an efficient implementation of SET-MLP, a sparse multilayer perceptron trained with the sparse evolutionary training procedure, is proposed to deal with high dimensional microarray datasets. This implementation makes use just of Python 3, sparse data structures from SciPy, and Cython. With this implementation, we have created for the first time in literature sparse MLP models with over one million neurons which can be trained on a standard laptop using a single CPU thread and without GPU. This is with two orders of magnitude more than state-of-the-art MLP models trained on commodity hardware.

Besides, we demonstrated on four microarray datasets with tens of thousands of input features and with up to just two thousands of samples that our approach reduces the number of connections quadratically in large MLPs (about 99.9 \% sparsity) while outperforming the state-of-the-art methods on these datasets for the classification task. Moreover, our proposed SET-MLP models showed to be robust to overfitting, imbalanced and data shift problems, which is not so usual for fully connected MLPs. Additionally, the results suggest that our proposed approach can cope efficiently with the ``curse of dimensionality", being capable of learning from small amounts of labeled data, and outperforming the state-of-the-art methods (ensembles of classifiers and feature selection methods) which are currently employed on high dimensional non-grid like data (or tabular data). 

In the future, we intend to put our emphasis on other types of neural layers, such as convolutional layers in CNN which have been widely used to deal with graphic data with grid-like topology. Furthermore, we intend to extend this work to address problems from other fields that suffer from the ``curse of dimensionality" and which have ultra high dimensional data (e.g. social networks, financial networks, semantic networks). The last but not the least future research direction would be to parallelize our implementation to use all CPU threads of a typical workstation efficiently and to incorporate it into usual Deep Learning frameworks, such as TensorFlow or PyTorch. This probably would allow us to scale with one order of magnitude more the SET-MLP models (up to the level of few tens of millions of neurons), while still using commodity hardware.

\section*{Acknowledgement}
We thank Ritchie Vink\footnote{\url{https://www.ritchievink.com/} Last visit 3rd June 2018.} for providing on Github.com a vanilla fully connected MLP implementation, and to
Thomas Hagebols\footnote{\url{https://github.com/ThomasHagebols} Last visit 25th Jan 2019.} for analyzing the performance of SciPy sparse matrix operations.

\bibliography{mybib}
\bibliographystyle{icml2019}

\end{document}